\documentclass[10pt,twocolumn,letterpaper]{article}

\usepackage{iccv}
\usepackage{times}
\usepackage{epsfig}
\usepackage{graphicx}
\usepackage{amsmath}
\usepackage{amssymb}

\usepackage{bm}

\usepackage{amsthm}
\usepackage{amsbsy}
\usepackage{booktabs}
\usepackage{enumitem}
\usepackage{mathtools}
\usepackage{subfigure}
\usepackage{tabularx}
\usepackage{multirow}
\usepackage{caption}
\usepackage{overpic}
\usepackage{xcolor}
\usepackage{comment}
\usepackage[title]{appendix}
\usepackage{balance}
\usepackage{makecell}
\usepackage[numbers,sort,compress]{natbib}
\usepackage[accsupp]{axessibility} 
\usepackage[bottom]{footmisc}

\usepackage{enumitem}
\setitemize{noitemsep,topsep=0pt,parsep=0pt,partopsep=0pt}

\DeclarePairedDelimiterX{\infdivx}[2]{(}{)}{%
  #1\;\delimsize\|\;#2%
}
\DeclarePairedDelimiterX{\inp}[2]{\langle}{\rangle}{#1, #2}

\DeclarePairedDelimiter{\norm}{\big\lVert}{\big\rVert}
\DeclarePairedDelimiter{\bignorm}{\Big\lVert}{\Big\rVert}
\DeclareMathOperator*{\argmin}{\arg\!\min}

\newlength{\oldparindent}
\newcommand{\modelname}{POP\xspace} 
\newcommand{\dataset}{ReSynth\xspace}

\renewcommand{\paragraph}[1]{\vspace{2pt}{\noindent\textbf{#1}}}

\newcommand*{\affaddr}[1]{#1} 
\newcommand*{\affmark}[1][*]{\textsuperscript{#1}}
\newcommand*{\email}[1]{\small{\texttt{#1}}}

\usepackage[pagebackref=true,breaklinks=true,letterpaper=true,colorlinks,bookmarks=false]{hyperref}

\iccvfinalcopy 

\def\iccvPaperID{1017} 

\begin{document}

\title{The Power of Points for Modeling Humans in Clothing}
\author{
Qianli Ma\affmark[1,2]\quad
Jinlong Yang\affmark[1] \quad
Siyu Tang\affmark[2]\quad
Michael J. Black\affmark[1]
\\
\affaddr{\affmark[1]Max Planck Institute for Intelligent Systems, T\"ubingen, Germany} \quad \affaddr{\affmark[2]ETH Z\"urich}\\
\email{\{qma,jyang,black\}@tuebingen.mpg.de},\quad \email{\{qianli.ma, siyu.tang\}@inf.ethz.ch}
}

\twocolumn[{%
\renewcommand\twocolumn[1][]{#1}%
\maketitle
\begin{center}
    \newcommand{\teaserwidth}{\textwidth}
\vspace{-0.2in}
   \centerline{
    \includegraphics[width=1\linewidth]{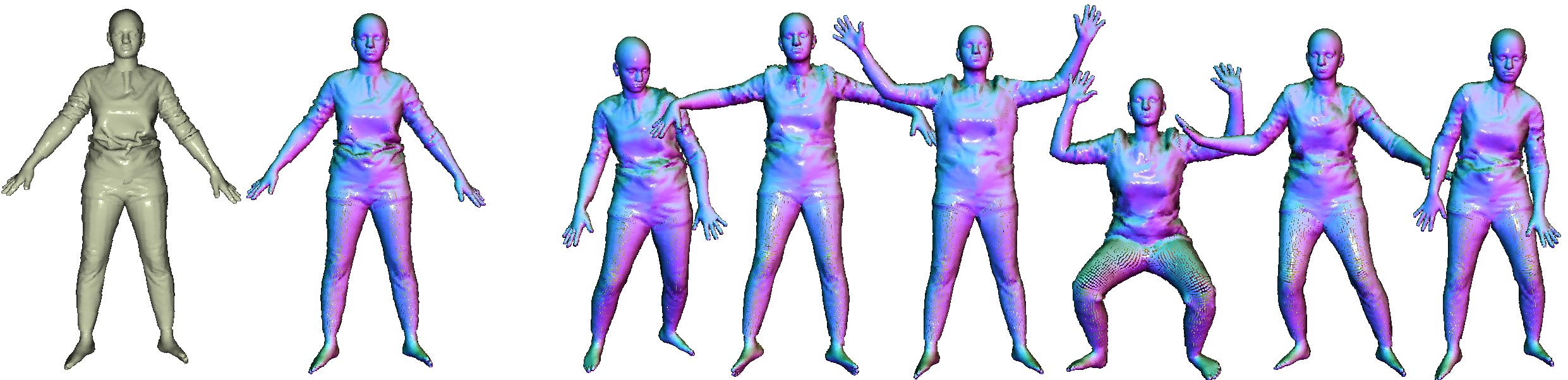}
    \put(-478, -10){\small{Unseen scan~\quad \ Fitted POP (point cloud) }}
    \put(-254, -10){\small{Animations with pose-dependent clothing deformation}}
     }
  \vspace{3pt}
  \captionof{figure}{The \textbf{Power of Points (POP)} model for clothed humans.
  Based on a novel articulated dense point cloud representation, our cross-outfit model, named POP, produces  pose-dependent shapes of clothed humans  with coherent global shape and expressive local garment details. 
  The trained model can be fitted to a \textit{single} scan of an unseen subject wearing an unseen outfit, and can animate it with realistic pose-dependent clothing deformations.
  The results are color-coded with predicted point normals and rendered with a simple surfel-based renderer.}
\label{fig:teaser}
\end{center}%
}]
\maketitle

\begin{abstract}
Currently it requires an artist to create 3D human avatars with realistic clothing that can move naturally.
Despite progress on 3D scanning and modeling of human bodies, there is still no technology that can easily turn a static scan into an animatable avatar.
Automating the creation of such avatars would enable many applications in games, social networking, animation, and AR/VR to name a few.
The key problem is one of representation.
Standard 3D meshes are widely used in modeling the minimally-clothed body but do not readily capture the complex topology of clothing.
Recent interest has shifted to implicit surface models for this task but they are computationally heavy and lack compatibility with existing 3D tools.
What is needed is a 3D representation that can capture varied topology at high resolution and that can be learned from data.
We argue that this representation has been with us all along --- the point cloud.
Point clouds have properties of both implicit and explicit representations that we exploit to model 3D garment geometry on a human body.
We train a neural network with a novel local clothing geometric feature to represent the shape of different outfits. 
The network is trained from 3D point clouds of many types of clothing, on many bodies, in many poses, and learns to model pose-dependent clothing deformations.
The geometry feature can be optimized to fit a previously unseen scan of a person in clothing, enabling the scan to be reposed realistically.
Our model demonstrates superior quantitative and qualitative results in both multi-outfit modeling and unseen outfit animation. 
The code is available for research purposes at {\small\url{https://qianlim.github.io/POP}}.

\end{abstract}

\section{Introduction}
Animatable clothed human avatars are required in many applications for 3D content generation.
To create avatars with naturally-deforming clothing, existing solutions either involve heavy artist work or require 4D scans for training machine-learning models~\cite{SCANimate:CVPR:2021}.
These solutions are expensive and often impractical. 
Instead, can we turn a single static 3D scan --- which can be acquired at low cost today even with hand-held devices --- into an animatable avatar?
Currently, no existing technology is able to do this and produce realistic clothing deformations.
Given a static scan, traditional automatic rigging-and-skinning methods ~\cite{baran2007automatic, feng2015avatar, liu2019neuroskinning} can be used to animate it, but are unable to produce pose-dependent clothing deformations. Physics-based simulations can produce realistic deformations, but require ``reverse-engineering'' a simulation-ready clothing mesh from the given scan.
This involves expert knowledge and is not fully automatic. 

Taking a data-driven approach, the goal would be to learn a model that can produce reasonable pose-dependent clothing deformation across different outfit types and styles and can generalize to unseen outfits. However, despite the recent progress in modeling clothed human body shape deformations~\cite{guan2012drape, patel20tailornet, lahner2018deepwrinkles, CAPE:CVPR:20, SCANimate:CVPR:2021,SCALE:CVPR:2021}, most existing models are \textit{outfit-specific} and thus cannot generalize to unseen outfits. 
To date, no such cross-garment model exists, due to several technical challenges. 

The first challenge lies in the choice of 3D shape representation. 
To handle outfits of different types and styles at once, the shape representation must handle changing topology, capture high-frequency details, be fast at inference time, and be easy to render.
Classical triangle meshes excel at rendering efficiency but are fundamentally limited by their fixed topology. 
The implicit surface representation is topologically flexible, but is in general computationally heavy and lacks compatibility with existing graphics tools.
Because point clouds are an explicit representation, they are efficient to render, but they can also be viewed as implicitly representing a surface.
Thus they are flexible in topology and, as the resolution of the point cloud increases, they can capture geometric details. 
While point clouds are not commonly applied to representing clothing, they are widely used to represent rigid objects and many methods exist to process them efficiently with neural networks~\cite{achlioptas2017learning,fan2017point,lin2018learning}.
In this work, we show that the seemingly old-fashioned point cloud is, in fact, a powerful representation for modeling clothed humans.

In recent work, SCALE~\cite{SCALE:CVPR:2021} demonstrates that a point cloud, grouped into local patches, can be exploited to represent clothed humans with various clothing styles, including those with thin structures and open surfaces.
However, the patch-based formulation in SCALE often suffers from artifacts such as  gaps between patches.
In this work, we propose a new shape representation of dense point clouds. 
For simplicity, we  avoid using patches, which have been widely used in recent point cloud shape representations~\cite{bednarik2020, deng2020better, deprelle2019learning, groueix2018atlasnet, SCALE:CVPR:2021}, and show that patches are not necessary. 
Instead, we introduce smooth local point features on a 2D manifold that regularize the points and enable arbitrarily dense up-sampling during inference.

Another challenging aspect of cross-outfit modeling concerns how outfits of different types and styles can be encoded in a single, unified, model. 
In most existing outfit-specific models, the model parameters (typically the weights of a trained shape decoder network) need to represent both the intrinsic, pose-\textit{in}dependent shape of a clothed person, and how this shape deforms as a function of the input pose. To factor the problem, we propose to isolate the intrinsic shape from the shape decoder by explicitly conditioning it with a \textit{geometric feature tensor}. The geometric feature tensor is learned in an auto-decoding fashion~\cite{park2019deepsdf}, with a constraint that a consistent intrinsic shape is shared across all examples of the same outfit. Consequently, the shape decoder can focus on modeling the pose-dependent effects and can leverage common deformation properties across outfits. 
At inference time, the geometric feature tensor can be optimized to fit to a scan of a clothed body with a previously unseen outfit, making it possible for the shape decoder to predict pose-dependent deformation of it based on the learned clothing deformation properties. 

These ideas lead to \modelname: our dense point cloud model that produces pose-dependent clothing geometry across different outfits and demonstrating the \textit{Power of Points} for modeling shapes of humans in clothing. 
\modelname is evaluated on both captured and synthetic datasets, showing state-of-the-art performance on clothing modeling and generalization to unseen outfits. 

In summary, our contributions are: (1) a novel dense point cloud shape representation with fine-grained local features that produces state-of-the-art detailed clothing shapes with various clothing styles;
(2) a novel geometry feature tensor that enables cross-garment modeling and generalization to unseen outfits; 
(3) an application of animating a static scan with reasonable pose-dependent deformations. 
The model and code are available for research purposes at {\small\url{https://qianlim.github.io/POP}}.
\section{Related Work}
\paragraph{Shape Representations for Clothed Human Modeling.}
Surface meshes are an efficient 3D representation that is compatible with graphics pipelines, and thus are the dominant choice for modeling clothing and clothed humans. With meshes, the clothing is represented either by deforming an unclothed body template~\cite{bhatnagar2019mgn,burov2021dsfn,CAPE:CVPR:20, Neophytou2014layered,tiwari20sizer,yang2018physics}, or using a separately defined template~\cite{guan2012drape, gundogdu2019garnet, lahner2018deepwrinkles,patel20tailornet,santesteban2019}.
While recent work successfully produces detailed geometry with graph convolutions~\cite{CAPE:CVPR:20}, multi-layer perceptrons (MLPs)~\cite{burov2021dsfn,patel20tailornet}, and UV map convolutions~\cite{jin2018pixel, lahner2018deepwrinkles}, meshes suffer from two fundamental limitations: the fixed topology and the requirement for template registration. This restricts their generalization to outfit styles beyond the pre-defined templates, and makes it difficult to obtain a common representation for various clothing categories.
Although recent work proposes adaptable templates~\cite{pan2019deep,zhu2020deep} that modify mesh connectivity, the need for registering training data to the mesh template remains challenging when a complex garment geometry is involved.

Neural implicit surfaces~\cite{chibane2020ndf, mescheder2019occupancy, park2019deepsdf}, on the other hand, do not require any pre-defined template, are flexible with surface topology, and have recently become a promising choice for reconstructing~\cite{Huang:ARCH:2020,he2021archpp, saito2019pifu, saito2020pifuhd, bozic2021neuraldeformationgraphs,yang2021s3,zheng2020pamir,Peng_2021_CVPR} and modeling~\cite{chen2021snarf,SMPLicit:2021,deng2019neural, SCANimate:CVPR:2021,LEAP:CVPR:21,palafox2021npm} shapes of 3D humans. Despite their ability to handle varied clothing topology, it remains an open challenge to realistically represent thin structures that are often present in daily clothing. 
Moreover, reconstructing an explicit surface from the implicit function costs cubic time with respect to the resolution, which restricts them from many practical applications. 

Point clouds are another classic 3D representation that supports arbitrary topology as well as thin structures. 
Going beyond prior work that generates a sparse point set~\cite{achlioptas2017learning,fan2017point,lin2018learning}, recent approaches~\cite{bednarik2020, deng2020better, deprelle2019learning, groueix2018atlasnet} use deep learning to generate structured point clouds with a set of surface patches. 
Leveraging the patch regularization, SCALE~\cite{SCALE:CVPR:2021} proposes an articulated dense point cloud representation to model clothing deformation.
However, the patch-based point clouds often suffer from overlap~\cite{groueix2018atlasnet} or separation~\cite{SCALE:CVPR:2021} between the patches, which degrades the geometric fidelity and visual quality.
Our model, \modelname, follows the spirit of these approaches to generate a dense, structured point cloud, but, in contrast to prior work, we deprecate the concept of patches and, instead, decode a dense point cloud from fine-grained local features. 
The resulting clothed human model shows high geometry fidelity, is robust to topology variations, and generalizes to various outfit styles.

\paragraph{Modeling Outfit Shape Space.} 
We categorize existing models for clothing or clothed bodies into three levels of increasing generalization capacity, as summarized in Tab.~\ref{tab:outfit_space_modeling}. Note that non-parametric models for clothed human reconstruction~\cite{saito2019pifu,saito2020pifuhd,zheng2020pamir} are out of the scope for this analysis.

\textit{Outfit-Specific}. Methods from this class need to train a separate model for every outfit instance~\cite{de2010stable,gundogdu2019garnet,lahner2018deepwrinkles,SCALE:CVPR:2021,Neophytou2014layered, santesteban2019,SCANimate:CVPR:2021,yang2018analyzing} or category (e.g.~all short-sleeve T-shirts)~\cite{guan2012drape,patel20tailornet,tiwari20sizer,wang2018learning}. For mesh-based methods in this category~\cite{de2010stable,guan2012drape,lahner2018deepwrinkles,Neophytou2014layered,patel20tailornet,tiwari20sizer,yang2018analyzing}, this characteristic stems from the need to manually define a mesh template: the fixed mesh topology fundamentally prohibits generalization to a different outfit category (e.g.~from  pants to a skirt).
These methods can, however, deal with size~\cite{guan2012drape,tiwari20sizer} or style~\cite{patel20tailornet, wang2018learning} within the template-defined category.  
Template-free methods~\cite{SCALE:CVPR:2021, SCANimate:CVPR:2021} require training a separate model for each outfit, i.e.~the intrinsic, pose-\textit{in}dependent shape information is stored in the model parameters; hence, the test-time generalization to an unseen outfit is restricted.

\textit{Multi-Outfit}. Combining multiple pre-defined mesh templates with a multi-head network or a garment classifier, the MGN model~\cite{bhatnagar2019mgn}, BCNet~\cite{jiang2020bcnet}, and DeepFashion3D~\cite{zhu2020deep} can reconstruct humans in a variety of clothing from images.
In a similar spirit, CAPE~\cite{CAPE:CVPR:20} uses a pre-labeled one-hot vector for outfit-type conditioning and can generate new clothing from four common categories with a single model. 
While training a single model for multiple outfits exploits the complementary information among training data, these methods do not show the ability to handle unseen garments beyond the pre-defined categories.

\textit{Arbitrary Outfit.} To overcome the limitations brought by the fixed topology of meshes, recent work opts for other representations that can unify different clothing categories and types. 
Shen et al.~\cite{shen2020gan} represent garments using 2D sewing pattern images that are applicable to arbitrary clothing categories. However, the final 3D garment shape is represented with a single manifold mesh that is not sufficiently expressive to represent the complexity and variety of real-world clothing.
Using neural implicit surfaces, SMPLicit~\cite{SMPLicit:2021} learns a topology-aware generative model for garments across multiple categories and shows continuous interpolation between them. However, the clothing geometry tends to be bulky and lacks details.
In contrast, our \modelname model
faithfully produces geometric details of various outfits, can generalize to unseen outfits, and demonstrates state-of-the-art performance on garment space modeling. 

\begin{table}[tb]
    \centering
    \caption{Data-driven models for clothing / clothed humans classified by garment space generalization.}
    \small
    \begin{tabular}{p{2.0cm}p{5.4cm}}
    \hline\hline
        \makecell[l]{Outfit-Specific} &  \makecell*[{{p{5.6cm}}}]{De~Aguiar~\cite{de2010stable}, DRAPE~\cite{guan2012drape}, GarNet~\cite{gundogdu2019garnet}, \\ 
        DeepWrinkles~\cite{lahner2018deepwrinkles}, SCALE~\cite{SCALE:CVPR:2021},\\
        Neophytou~\cite{Neophytou2014layered}, TailorNet~\cite{patel20tailornet},\\ 
         SCANimate~\cite{SCANimate:CVPR:2021}, Santesteban~\cite{santesteban2019}, \\
         Sizer~\cite{tiwari20sizer}, Wang~\cite{wang2018learning}, Yang~\cite{yang2018analyzing}.
         }\\
        \hline
        \makecell[l]{Multi-Outfit} & \makecell[l]{MGN~\cite{bhatnagar2019mgn}, BCNet~\cite{jiang2020bcnet}, CAPE~\cite{CAPE:CVPR:20},\\
        Vidaurre~\cite{vidaurre2020fully},  DeepFashion3D~\cite{zhu2020deep}.
        }\\
        \hline
        \makecell[l]{Arbitrary Outfit} & SMPLicit~\cite{SMPLicit:2021}, Shen~\cite{shen2020gan}, \textbf{\modelname (Ours)}.\\
    \hline\hline
    \end{tabular}
    \label{tab:outfit_space_modeling}
    \vspace{-10pt}
\end{table}
\section{Method}
Our goal is to learn a single, unified, model of high-fidelity pose-dependent clothing deformation on human bodies across multiple outfits and subjects. 
We first introduce an expressive point-based representation that preserves geometric details and flexibly models varied topology (Sec.~\ref{sec:representaion}). 
Using this, we build a cross-outfit model enabled by a novel geometric feature tensor (Sec.~\ref{sec:clo_modeling}).

\begin{figure*}[t]
    \centering
    \includegraphics[width=\textwidth]{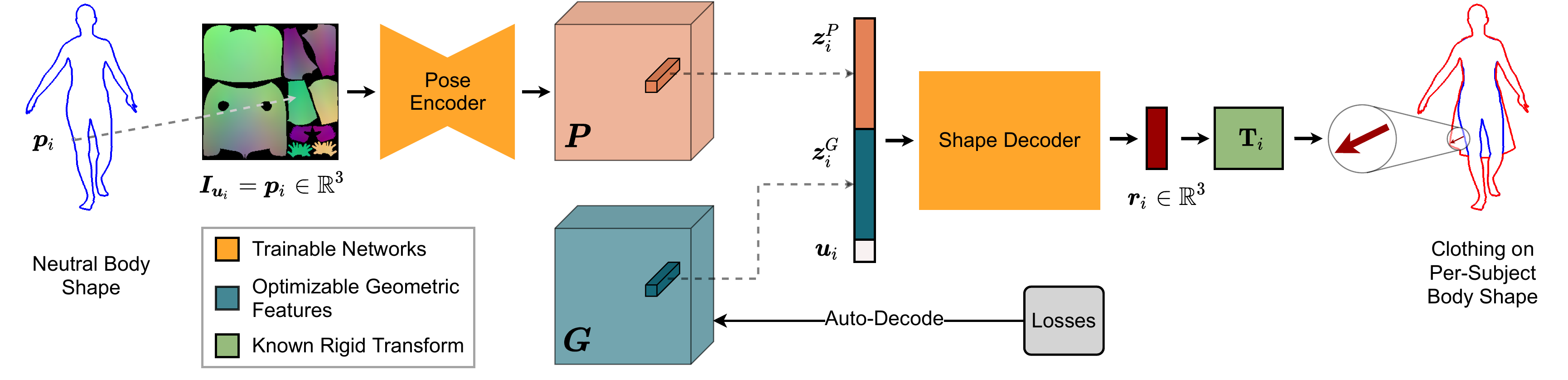}
    \caption{\small \textbf{Overview of \modelname.} Given a posed but unclothed body model (visualized as a blue 2D contour), we record the 3D positions $\bm{p}_i$ of its surface points on a UV positional map $\bm{I}$, and encode this into a pose feature tensor $\bm{P}$. A garment geometry feature tensor $\bm{G}$ is an optimizable variable that is pixel-aligned with $\bm{P}$, and learned per-outfit in an auto-decoder manner.
    The 2D image-plane coordinate $\bm{u}_i$ describes the relative location of the points on the body surface manifold. The shape decoder queries these locations and predicts displacement vectors $\bm{r}_i$ based on the points' local pose and geometry features.
    }
    \label{fig:method_overview}
    \vspace{-10pt}
\end{figure*}

As illustrated in Fig.~\ref{fig:method_overview}, given an unclothed body, the model outputs the 3D clothed body by predicting a displacement field from the body surface based on local pose and geometric features. 
The trained model can be fitted to a scan of a person in previously unseen outfits and this scan can be animated with pose-dependent deformations (Sec.~\ref{sec:losses}).

\subsection{Representing Humans with Point Clouds}\label{sec:representaion}
Point-based representations~\cite{groueix20183d,SCALE:CVPR:2021} possess topological flexibility and fast inference speed, giving them an advantage over meshes or implicit functions for modeling articulated humans.
In this work, we formulate a structured point cloud representation for modeling 3D clothed humans 
by learning a mapping from a 2D manifold to a 3D deformation field, in a similar form to AtlasNet~\cite{groueix2018atlasnet}:
\begin{equation}
\label{eq:pop_math}
\bm{r}_i = f_\mathbf{w}(\bm{u}_i; \bm{z}_i): \mathbb{R}^2 \times \mathbb{R}^\mathcal{Z} \to \mathbb{R}^3,
\end{equation}
where $\bm{r}_i$ is a displacement vector, $f_\mathbf{w}(\cdot)$ is a multi-layer perceptron (MLP) with weights $\mathbf{w}$, $\bm{u}_i$ is a 2D parameterization of a point $i$ that denotes its relative location on the body surface, and $\bm{z}_i$ is the point's local feature code containing the shape information. 

We deviate from other recent point-based human representations in two key ways: 
1) We use fine-grained \textit{per-point local features} $\bm{z}_i$ as opposed to a single global feature~\cite{groueix20183d,groueix2018atlasnet} or per-patch features~\cite{SCALE:CVPR:2021} in prior work.
2) We predict the \textit{clothing deformation field} on top of the canonically posed body, instead of absolute Euclidean point coordinates~\cite{groueix20183d} or local patch deformations~\cite{SCALE:CVPR:2021}. Both design choices lead to significant improvements in representation power, as detailed below.

\paragraph{Continuous Local Point Features.} 
Recent work shows the advantage of using a local latent shape code over a global code: both for neural implicit surfaces~\cite{chabra2020deep,genova2020local,jiang2020local,Peng2020ECCV} and point clouds~\cite{SCALE:CVPR:2021}.
Decoding shape from local codes significantly improves geometry quality.
In particular for modeling humans, SCALE~\cite{SCALE:CVPR:2021} successfully leverages local features to represent pose-aware garment shape and demonstrates a significant qualitative improvement against prior work that uses a single global feature code~\cite{groueix20183d}.

\begin{figure}[tb]
    \centering
    \includegraphics[width=\linewidth]{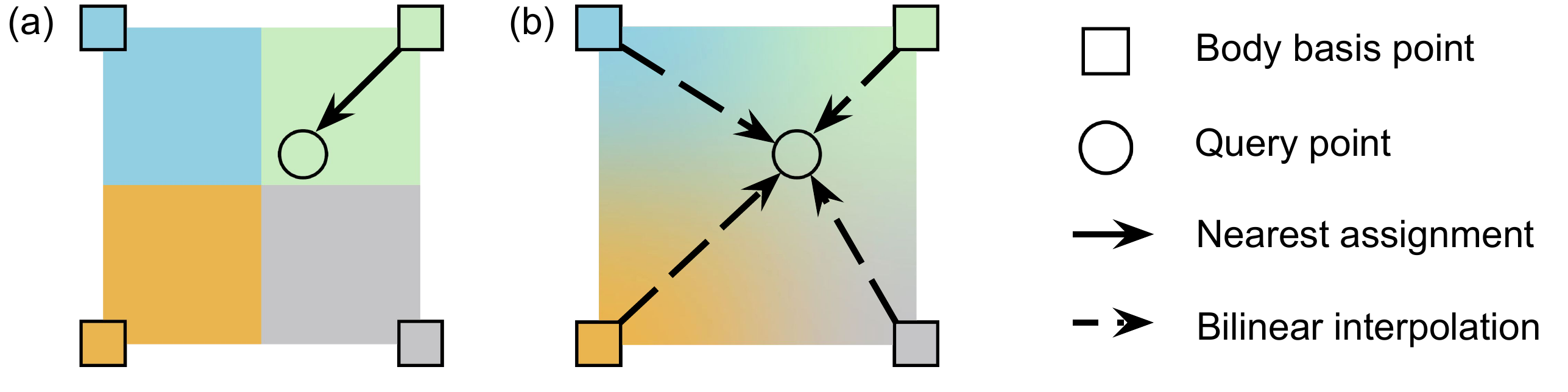}
    \caption{\small 2D illustration of the point feature assignment for a region on the body manifold between the body basis points. Colors represent features. \textbf{(a)} The nearest neighbor assignment used by SCALE~\cite{SCALE:CVPR:2021},  causes ``patchy'' predictions. \textbf{(b)} Our bilinear feature interpolation results in smoothly varying features on the manifold. }
    \label{fig:blerp_feat}
    \vspace{-1em}
\end{figure}

However, the local feature in SCALE is still discretely distributed on a set of pre-defined basis points $\bm{u}_i$ on the body manifold. Each feature code is decoded into multiple points in a neighborhood (a ``patch'') in the output space, but it varies discretely across patch boundaries. 
This is equivalent to the nearest neighbor (on the body manifold) assignment of the features, Fig.~\ref{fig:blerp_feat}(a).
This discrete characteristic limits the quality of the geometry. 
As shown in Fig.~3 of \cite{SCALE:CVPR:2021} and our Sec.~\ref{exp:ours_vs_baselines}, the patches are typically isolated from each other, leading to uneven point distributions, hence poor mesh reconstruction quality. 

To address this problem, we make the local features more fine-grained in two ways. First, we define a denser set of basis points $\bm{u}_i$, together with their local features, on the body manifold. In practice, this amounts to simply increasing the resolution of the body UV map (see Sec.~\ref{sec:clo_modeling}).
Second, we further diffuse the feature over the body surface: for a query coordinate on the body surface, we compute its feature by bilinearly interpolating the features from its 4 nearest basis points, Fig.~\ref{fig:blerp_feat}. 
As a result, the network output can be evaluated at arbitrarily high resolution by querying the decoder $f_\mathbf{w}(\cdot)$ with any point on the body surface, which we also denote as $\bm{u}_i$ from now on with a slight abuse of notation.

\paragraph{Local Transformations.}
The clothing deformation vector $\bm{r}_i$ in Eq.~\eqref{eq:pop_math} is predicted on top of the body in the \textit{canonical pose}. 
Thus a large portion of shape variation is explained by body articulation so that the network can focus on modeling the residual shape.
Note that unlike the mesh vertex offset representation of clothing~\cite{bhatnagar2019mgn,CAPE:CVPR:20,pons2017clothcap}, our formulation does not assume a constant topology, and can thus model various outfit styles such as pants and dresses.

To reconstruct the clothed body in the posed space, we transform $\bm{r}_i$ according to the transformation matrix $\mathbf{T}_i$ at $\bm{u}_i$ that is given 
by the fitted body model. The point position from the clothed body in the posed space is then given by:
\begin{equation}\label{eq:local2world}
    \mathbf{x}_i = \mathbf{T}_i \cdot \bm{r}_i + \bm{p}_i,
\end{equation}
where $\bm{p}_i$ is the 3D Euclidean coordinate of $\bm{u}_i$ on the posed unclothed body.
Note that we branch out the final layers of $f_\mathbf{w}(\cdot)$ such that it also predicts the normal $\bm{n}(\mathbf{x}_i)$ of each point, which is transformed with the rotation part of $\mathbf{T}_i$. 

Since our local point features are continuous over the body surface, we also perform barycentric interpolation to obtain continuously varying $\mathbf{T}_i$, in the same way as LoopReg~\cite{bhatnagar2020loopreg}.
\subsection{Cross-garment Modeling with a Single Model}\label{sec:clo_modeling}

With our structured point cloud representation, we now build a system that models pose-dependent deformations of various garments, from different categories, of different shapes and topology, dressed on different body shapes, using a \textit{single} model. 
This is achieved by introducing a novel geometric feature tensor. Practically, we decompose the local feature $\bm{z}_i$ in Eq.~\eqref{eq:pop_math} into pose $\bm{z}_i^P$ and garment geometry $\bm{z}_i^G$, as illustrated in Fig.~\ref{fig:method_overview} and detailed below.

\paragraph{Body Shape Agnostic Pose Encoding.}
We first condition our network with learned local body pose features $\bm{z}_i^P$ such that the output garment deformation is pose-aware. 
We adopt the approach based on the UV positional map of the posed body as used in~\cite{SCALE:CVPR:2021} as it shows better pose generalization than the traditional pose parameter conditioning~\cite{CAPE:CVPR:20,patel20tailornet, yang2018analyzing}.
As shown in Fig.~\ref{fig:method_overview}, 
a UV positional map $\bm{I}\in\mathbb{R}^{H\times W\times 3}$ is a 2D parameterization of the body manifold, where each valid pixel corresponds to a point on the posed body surface. 
The 2D image-plane coordinate of a pixel describes its manifold position: $\bm{u}_i=(u, v)_i$. The pixel value records the point's location in $\mathbb{R}^3$: $\bm{p}_i=\bm{I}_{\bm{u}_i}$.
A UNet~\cite{ronneberger2015u} encodes $\bm{I}$ into a pose feature tensor $\bm{P}\in \mathbb{R}^{H\times W\times 64}$, 
where $H, W$ are the spatial dimensions of the feature tensor, and each ``pixel'' from $\bm{P}$ is a 64-dimensional pose feature code: $\bm{z}_i^P = \bm{P}_{\bm{u}_i}\in \mathbb{R}^{64}$. Because of the receptive field of the UNet, the learned pose features can contain global pose information from a larger neighborhood when necessary.

The UV positional map naturally contains information about body shape. 
To generalize to different subjects, we use posed bodies of a \textit{neutral shape} for the pose encoding.
Still, the predicted clothing deformations are added to each subject's body respectively. 

\paragraph{Geometric Feature Tensor.} 
In most learning-based, outfit/subject-specific clothed body models~\cite{SCALE:CVPR:2021,CAPE:CVPR:20,SCANimate:CVPR:2021, yang2018analyzing}, the clothing shape information is contained in the parameters of the trained shape decoder network, limiting  generalization to unseen outfits. 
What is needed for cross-outfit modeling is a mechanism that decouples the intrinsic, pose-\textit{in}dependent shape of a clothed person from the decoder, so that it can focus on modeling \textit{how} the shape deforms with the pose. To that end, we propose to explicitly condition the shape decoder with a \textit{geometric feature tensor} $\bm{G}\in \mathbb{R}^{H\times W\times 64}$, Fig.~\ref{fig:method_overview}.

The geometric feature tensor follows the spirit of being local and is pixel-aligned with the pose feature tensor. Each of its ``pixels'' represents a local shape feature on a body point: $\bm{z}_i^G = \bm{G}_{\bm{u}_i}\in$ $\mathbb{R}^{64}$. Unlike the pose features, the geometry features are learned in an auto-decoding~\cite{park2019deepsdf} fashion; i.e.~they are updated during training such that the optimal representation for the garment geometry is discovered by the network itself. 
Importantly, we use a consistent $\bm{G}$ for each outfit across all of its training examples (in different poses).
In this way, $\bm{G}$ is enforced to be a pose-agnostic canonical representation of each outfit's geometry. 

Our geometry feature tensor plays a similar role as the pre-defined clothing templates in many mesh-based garment models~\cite{gundogdu2019garnet,patel20tailornet,santesteban2019,vidaurre2020fully}. However, by auto-decoding the geometric features, we get rid of the reliance on manual template definition.
More importantly, our decoder network and the neural geometry feature are fully differentiable. This enables generalization to \textit{unseen} garments at test time, which is done by optimizing $\bm{G}$ to fit the target clothed body scan. See Sec.~\ref{sec:losses} for details.

\paragraph{Shared Garment Shape Decoder.}
With the introduced local pose and geometry features, we can re-write Eq.~\eqref{eq:pop_math} in a more concrete form: 
$\bm{r}_i = f_\mathbf{w}([\bm{u}_i,\bm{z}_i^P, \bm{z}_i^G])$, 
where $[\cdot,\cdot,\cdot]$ denotes concatenation. 
While $\bm{z}_i^G$ is optimized for each garment and $\bm{z}_i^P$ is acquired from each pose, $f_\mathbf{w}(\cdot)$ is shared for the entire set of all garments and poses. By training on many outfits and poses, the decoder learns common properties of clothing deformation, with which it can animate scans in unseen outfits at test-time, as described below. 
\subsection{Training and Inference}\label{sec:losses}
\paragraph{Loss Functions.}
We train \modelname with multi-subject and outfit data. During training, the parameters of the UNet pose encoder, the garment shape decoder, and the geometric feature tensor $\bm{G}$ are optimized, by minimizing the loss function:
\begin{equation}\label{eq:total_loss}
\mathcal{L}_\textrm{total} = \lambda_\textrm{d}\mathcal{L}_\textrm{d}+ \lambda_\textrm{n}\mathcal{L}_\textrm{n}  + \lambda_\textrm{rd}\mathcal{L}_\textrm{rd} + \lambda_\textrm{rg}\mathcal{L}_\textrm{rg},
\end{equation}
where the $\lambda$'s are weights that balance the loss terms, and the $\mathcal{L}$'s are the following loss terms.

First, the normalized Chamfer Distance $\mathcal{L}_\textrm{d}$ is employed to penalize the average bi-directional point-to-point $L2$ distances between the generated point cloud $\mathbf{X}$ and a sampled point set $\mathbf{Y}$ from the ground truth surface: $\mathcal{L}_\textrm{d} = d(\mathbf{x},\mathbf{y}) = $
\begin{equation}
\begin{aligned}\label{eq:chamfer}
\frac{1}{M} \sum_{i=1}^{M} \min _{j}\norm{\mathbf{x}_{i}-\mathbf{y}_j}_2^{2} +\frac{1}{N}\sum_{j=1}^N \min_{i}\norm{\mathbf{x}_{i}-\mathbf{y}_j}_2^{2},
\end{aligned}
\end{equation}
where $M,N$ are the number of points from the generated point cloud and the ground truth surface, respectively.

The normal loss $\mathcal{L}_\textrm{n}$ is the averaged $L1$ discrepancy between the normal prediction on each generated point and its nearest neighbor from the ground truth point set:
\begin{equation}\label{eq:normal_loss}
\mathcal{L}_\textrm{n} = \frac{1}{M}\sum_{i=1}^M\bignorm{\bm{n}(\mathbf{x}_{i})
    - \bm{n}(\argmin_{\mathbf{y}_j\in\mathbf{Y}}d(\mathbf{x}_{i},\mathbf{y}_j))}_1,
\end{equation}
where $\bm{n}(\cdot)$ denotes the unit normal of the given point. 

An $L2$ regularizer $\mathcal{L}_\textrm{rd}$ discourages the predicted point displacements from being extremely large. Similarly, the term $\mathcal{L}_\textrm{rg}$ penalizes the $L2$-norm of the vectorized geometric feature tensor to regularize the garment shape space:
\begin{equation}\label{eq:rgl}
\mathcal{L}_\textrm{rd} = \frac{1}{M}\sum_{i=1}^{M}\norm{\mathbf{r}_{i}}_2^2,\qquad
\mathcal{L}_\textrm{rg} = \frac{1}{C}\sum_{m=1}^{C}\norm{\bm{G}_{m}}_2^2,
\end{equation}
where $C$ is the number of garments seen in training.

The detailed model architecture, hyper-parameters and training procedure are provided in the SupMat.

\paragraph{Inference: Scan Animation.} At test-time, \modelname can generalize to unseen poses of both the previously seen and unseen outfits. For a seen outfit, we use its geometric feature tensor optimized from training, and infer the clothing deformation on unseen poses with a simple forward pass of the network.

To test on a scan $\hat{\mathbf{Y}}$ of a human wearing unseen clothing, we first fix the weights of the UNet pose encoder and the shape decoder $g_w(\cdot)$, and optimize the geometric feature tensor such that the total loss against the scan is minimized:
\begin{equation}\label{eq:test_unseen}
\hat{\bm{G}} = \argmin \mathcal{L}_\textrm{total}(\hat{\mathbf{Y}}).
\end{equation}
Afterwards, the estimated $\hat{\bm{G}}$ is  fixed, and is then treated as in the case of a seen garment.

As with other point-based human models, the point cloud generated by \modelname can either be meshed using classical tools such as the Poisson Surface Reconstruction (PSR)~\cite{kazhdan2006poisson}, or directly rendered into realistic images using recent point-based neural rendering techniques~\cite{aliev2019neural,kolos2020transpr,prokudin2020smplpix}.
However, in this work, we do not rely on neural rendering to inpaint the gaps between the points. Instead, we show qualitative results using a simple surfel-based renderer to more directly highlight the geometric properties of the smooth, high-resolution, human point cloud generated by \modelname.

\section{Experiments}
Due to the lack of comparable existing work on POP's two key features, namely cross-outfit modeling and single scan animation, we first evaluate its representation power on a simpler but related task: outfit-specific shape modeling, and compare with two state-of-the-art methods (Sec.~\ref{exp:ours_vs_baselines}). 
We then discuss the efficacy of our cross-outfit learning formulation (Sec.~\ref{exp:cross_outfit}) and demonstrate single scan animation (Sec.~\ref{exp:scan_animate}). 
\begin{table*}[t]
    \centering
    \caption{\label{tab:exp_summary} Results of pose-dependent deformation prediction on unseen test sequences from the captured CAPE dataset and our \dataset data. Best results are in \textbf{boldface}.}
    \small
    \begin{tabular}{lcccccc|cccccc}
    \hline\hline
    \multirow{4}{*}{Methods}& \multicolumn{6}{c}{CAPE Data}  & \multicolumn{6}{c}{\dataset Data} \\
    & \multicolumn{3}{c}{Chamfer-$L_2~(\times 10^{-4}m^2) \downarrow$} & \multicolumn{3}{c}{Normal diff. $(\times 10^{-1}) \downarrow$} & \multicolumn{3}{c}{Chamfer-$L_2~(\times 10^{-4}m^2) \downarrow$} & \multicolumn{3}{c}{Normal diff. $(\times 10^{-1}) \downarrow$}\\
    \cline{2-13}
    & \multirow{2}{*}{Mean} & Outfit & Outfit & \multirow{2}{*}{Mean} & Outfit & Outfit & \multirow{2}{*}{Mean} & Outfit & Outfit & \multirow{2}{*}{Mean} & Outfit & Outfit\\
    & & Median & Max &  & Median & Max &  & Median & Max &  & Median & Max \\
    \hline
    NASA~\cite{deng2019neural} & 6.087 & 1.190 & 32.35 & 1.275 & 1.277 & 1.497 & -- & -- & -- & -- & -- & -- \\
    SCALE~\cite{SCALE:CVPR:2021} & 0.721 & 0.689 & 0.971 & 1.168  & 1.170 & 1.335 & 1.491 & 0.680 & 8.451 & 1.041 & 1.054 & 1.321 \\
    Ours, per-outfit & 0.639 & 0.607 & 0.831 & 1.146 & 1.150 & 1.293 & \textbf{1.356} & 0.651 & \textbf{7.339} & \textbf{1.013} & \textbf{1.006} & 1.289 \\
    Ours, multi & \textbf{0.592} & \textbf{0.550} & \textbf{0.757} & \textbf{1.115} & \textbf{1.116} & \textbf{1.256} & 1.366 & \textbf{0.635} & 7.386 & 1.022 & 1.037 & \textbf{1.280}\\
    \hline
    $1/2$ Data  & 0.598 & 0.560 & 0.765 & 1.122 & 1.127 & 1.257 & 1.405 & 0.665 & 7.414 & 1.032 & 1.042 & 1.299 \\
    $1/4$ Data  & 0.621 & 0.586 & 0.841 & 1.134 & 1.142 & 1.271 & 1.406 & 0.674 & 7.469 & 1.032 & 1.043 & 1.296 \\
    $1/8$ Data  & 0.662 & 0.623 & 0.992 & 1.165 & 1.176 & 1.310 & 1.490 & 0.720 & 7.859 & 1.050 & 1.056 & 1.326 \\
     \hline\hline
    \end{tabular}
    \vspace{-1.6em}
\end{table*}

\paragraph{Datasets.}
We train and evaluate our method and baselines on both a captured clothed human dataset, CAPE~\cite{CAPE:CVPR:20}, and our new synthetic dataset called \textit{\dataset}.
From the CAPE dataset, we use the three subjects (00096, 00215, 03375) that contain the most abundant outfit variations (14 outfits in total) to compare the representation capacity of different methods.
The synthetic \dataset dataset is created with a larger variation in outfit shapes, styles, and poses.
We worked with a professional clothing designer to create 3D outfit designs that faithfully reflect those in a set of commercial 3D clothed human scans (Renderpeople~\cite{renderpeople}), resulting in 24 outfits including challenging cases such as skirts and jackets.
We then use physics simulation to drape the clothing on the 3D bodies from the AGORA dataset \cite{AGORA:CVPR:21}, which we animate to generate many poses.
Details of the datasets are provided in the SupMat., and we will release \dataset for research purposes.

\paragraph{Baselines.}
To evaluate the representation power of our model, we first compare with two recent methods for pose-dependent human shape modeling (Sec.~\ref{exp:ours_vs_baselines}): NASA~\cite{deng2019neural} and SCALE~\cite{SCALE:CVPR:2021}.
To evaluate the effectiveness of our cross-outfit modeling formulation (Sec.~\ref{exp:cross_outfit}), we compare two versions of our model: per-outfit trained and a cross-outfit model trained with data from all outfits, both using the same architecture.
For animating unseen scans (Sec.~\ref{exp:scan_animate}), we qualitatively compare with classical Linear Blend Skinning using the SMPL~\cite{loper2015smpl} body model.

\paragraph{Metrics.}
We quantitatively evaluate each method using the Chamfer Distance (in $m^2$, Eq.~\eqref{eq:chamfer}) and the $L1$ normal discrepancy (Eq.~\eqref{eq:normal_loss}), computed over the 50K points generated by our method and SCALE. For the implicit surface baseline NASA, the points for evaluation are sampled from surface extracted using Marching Cubes~\cite{lorensen1987marching}.
To evaluate the visual quality of the generated results, we perform a large-scale user study on the Amazon Mechanical Turk (AMT) and report the percentage of users that favor the results from our method over the baseline. Details of the user study are provided in the SupMat.

\subsection{Representation Power}\label{exp:ours_vs_baselines}
Tab.~\ref{tab:exp_summary} summarizes the numerical results of reconstructing pose-dependent garment shape from different methods, tested with seen outfits on \textit{unseen} motion sequences. 
As the difficulty of shape modeling varies greatly across different outfit types (e.g.~how a loose jacket deforms is much more complex than that of a tight T-shirt), we report three types of statistics to holistically reflect the performance of each model: the mean error averaged across all test examples from all outfits, the median of the per-outfit calculated mean error (denoted as ``Outfit Median'' in Tab.~\ref{tab:exp_summary}), and the per-outfit maximum error (``Outfit Max'').

\paragraph{Comparison with SoTA.}
The upper section of Tab.~\ref{tab:exp_summary} shows a comparison with NASA~\cite{deng2019neural} and SCALE~\cite{SCALE:CVPR:2021}. 
NASA represents the body shape with an ensemble of articulated occupancy functions defined per body part.
As it requires pre-computing occupancy values for training, it is in general not applicable to our synthetic data, which typically does not contain water-tight meshes.  
SCALE produces a point cloud of clothed bodies, where the points are grouped into local patches.
Notably, both methods need to train a separate model per outfit. 
Under the same outfit-specific training setting, \modelname outperforms both baselines on both datasets under all metrics.
We further conduct an AMT user study to evaluate the perceptual quality, comparing \modelname side-by-side against the strongest baseline, SCALE. 
On the CAPE data, 89.8\% participants rate \modelname's results as having ``higher visual quality'' than SCALE (10.2\%); while on \dataset, 84.8\% users favor \modelname over SCALE (15.2\%).

\begin{figure}[tb]
    \centering
    \includegraphics[width=\linewidth]{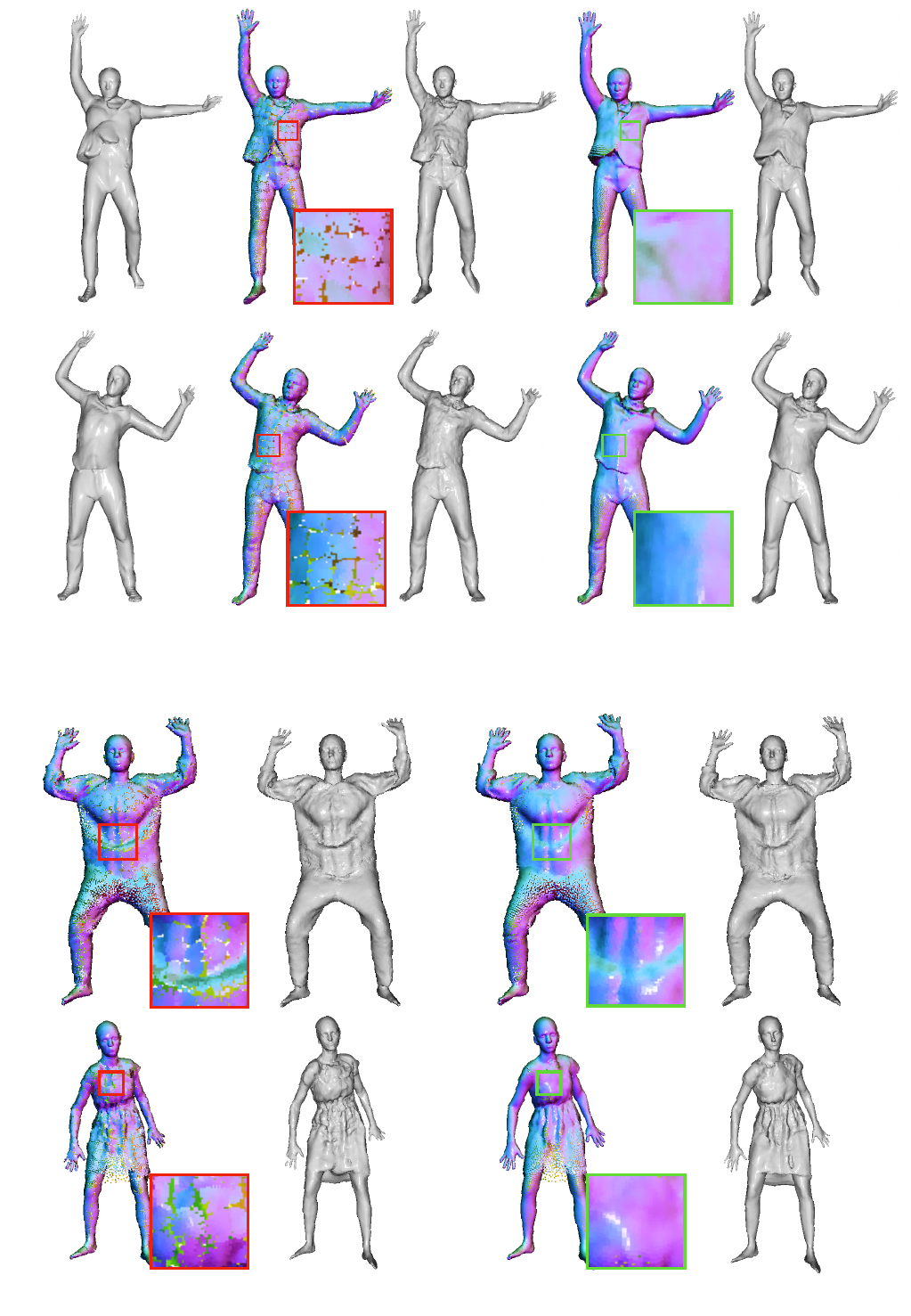}
    \put(-230,167){\footnotesize{NASA~\cite{deng2019neural} \quad SCALE~\cite{SCALE:CVPR:2021} \quad Meshed \qquad \quad Ours \qquad \quad Meshed}}
    \put(-226,-1){\footnotesize{SCALE~\cite{SCALE:CVPR:2021} \qquad ~~ Meshed \qquad \quad~~ \; Ours \qquad \qquad \; Meshed}}
    \caption{\small Comparison with SoTAs on the CAPE (upper 2 rows) and \dataset (lower 2 rows) data. 
    The dense point clouds from \modelname (cross-outfit model) are clean and preserve local details, while baseline methods suffer from artifacts. Note the point clouds are colored according to predicted normal and rendered with surfels~\cite{pfister2000surfels}. Best viewed zoomed-in on a color screen.}
    \vspace{-25pt}
    \label{fig:vs_sotas}
\end{figure}

The differences in perceptual quality can be seen in Fig.~\ref{fig:vs_sotas}. 
All approaches provide clear pose-dependent effects. 
However, NASA suffers from non-smooth transitions between separately-modeled body parts and is unable to handle thin structures and open surfaces such as the skirt in \dataset. 
SCALE, on the other hand, produces a smooth global shape with local details, but the isolation between the patches leads to sub-optimal visual quality.
In contrast, the dense point clouds generated by \modelname manifest a coherent overall shape and expressive local details. This illustrates the advantage of our continuous local point features as opposed to the discretely defined patch features.

\subsection{Cross-Outfit Modeling}\label{exp:cross_outfit}
A key feature of \modelname is the ability to handle multiple outfits of varied types using a single model, without pre-defining garment templates. 
As shown in Tab.~\ref{tab:exp_summary}, our cross-outfit model trained with all subjects and outfits (``Ours, multi'') reaches overall comparable performance to the outfit-specific models (``Ours, per-outfit'') on the synthetic \dataset data, and has significantly lower error on the CAPE dataset. Still, on both datasets, the cross-outfit model consistently outperforms the two baselines analyzed above.

The result on the CAPE data reveals the advantage of cross-outfit modeling: the information between different but similar outfit styles can be shared. The CAPE dataset mostly consists of simple and similar garments such as long/short T-shirts and pants, but the motions are performed differently by different subjects. Intuitively, training a cross-outfit model leads to mutual data augmentation in the pose space among different outfits, hence a better generalization to unseen poses as seen in the numerical results. 
In contrast, the synthetic data are simulated with a consistent set of motion sequences for all outfit types with largely varied geometry. As a result, the inter-outfit knowledge sharing is limited, leading to similar performance between per-outfit and cross-outfit models.

\paragraph{Robustness Against Limited Data.}
In the lower section of Tab.~\ref{tab:exp_summary}, we evaluate the performance of our cross-outfit model trained with subsets sampled from the full training set.
Even with only $1/4$ of data from each outfit, our cross-outfit model is comparable to the outfit-specific models of SCALE that are trained with full data. 

\subsection{Single Scan Animation}\label{exp:scan_animate}
\begin{figure}[tb]
    \centering
    \includegraphics[width=0.95\linewidth]{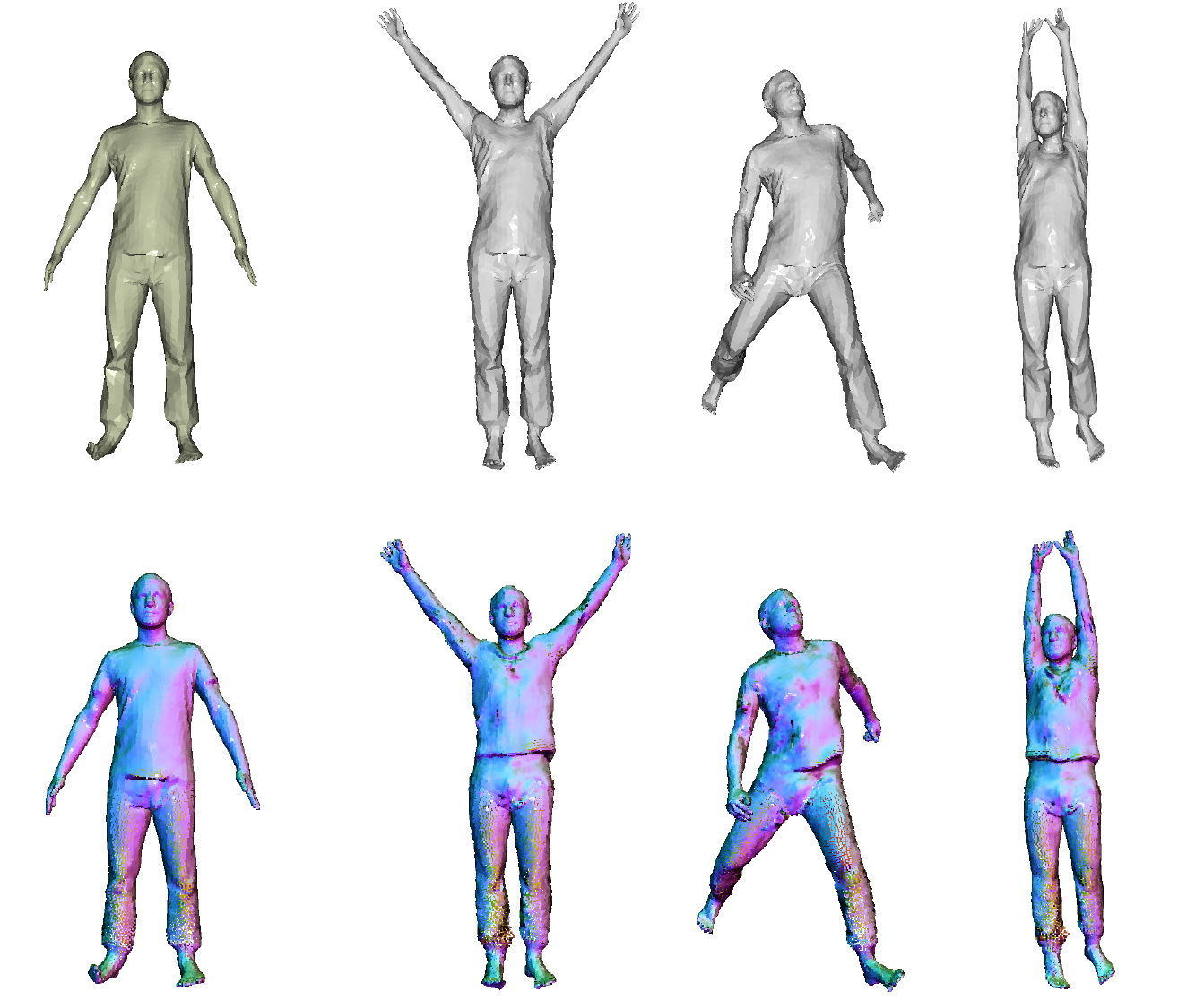}
    \put(-205,90){\footnotesize{Scan \qquad \qquad \qquad \qquad Animations with LBS}}
    \put(-213,-8){\footnotesize{Fitted POP \qquad \qquad \qquad \quad Animations with POP}}
    \caption{\small Comparison of animation with \modelname and an LBS-based method. The unseen scan from the CAPE dataset is animated on unseen motions.}
\vspace{-1em}
    \label{fig:repose_unseen_cape}
\end{figure}

\begin{figure}[tb]
    \centering
    \includegraphics[width=0.95\linewidth]{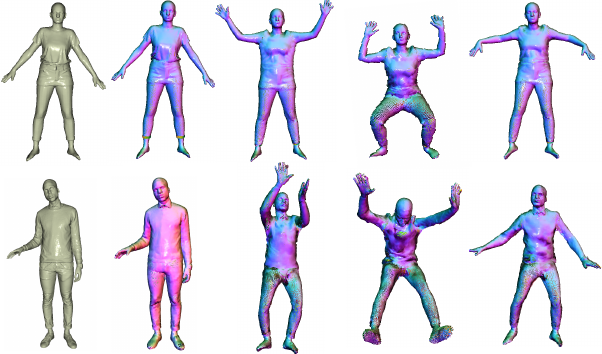}
    \put(-213,-8){\footnotesize{Scan \qquad Fitted POP \qquad \qquad Animations with POP}}
    \caption{\small Animation of an unseen example from \dataset (upper row) and a real-world capture (lower row) with unseen motions.}
    \vspace{-1.2em}
    \label{fig:repose_unseen_rp}
\end{figure}

Once trained, our cross-outfit \modelname model can be fit to a single scan of an unseen subject with an unseen outfit, and then animated with pose-dependent clothing variation, as described in Sec.~\ref{sec:losses}. Notably, this is a very challenging task since it requires generalization in both the outfit shape space and pose space.
Figure \ref{fig:repose_unseen_cape} qualitatively compares our model (trained on CAPE data) and the classical Linear Blend Skinning (LBS) technique that uses the SMPL model~\cite{loper2015smpl} to animate the given unseen scan with an unseen motion. 
Here we use sampled points from the mesh provided in the CAPE dataset as the target scan. As  LBS uses simple rigid transformations of the body parts only, it cannot produce complex pose-dependent shape variation. 
In contrast, \modelname produces reasonable and vivid clothing deformation such as the lifted hems.
In Fig.~\ref{fig:repose_unseen_rp} we deploy \modelname (trained on the \dataset data) to animate an unseen scan from \dataset and one from real-world captures~\cite{renderpeople}, respectively.
Note that the latter test is much more challenging due to the domain gap between our synthetic training data and the captured test data. 
\modelname produces high-quality clothing geometry on the \dataset test example and generates reasonable animation for the challenging real-world scan, which opens the promising new direction of automatic 3D avatar creation from a ``one-shot'' observation.

\section{Conclusion}
We have introduced \modelname, a dense, structured point cloud shape representation for articulated humans in clothing.
Being template-free, geometrically expressive, and topologically flexible, \modelname models clothed humans in various outfit styles with a single model, producing high-quality details and realistic pose-dependent deformation. The learned \modelname model can also be used to animate a single human scan from an unseen subject and clothing.
Our evaluations on both captured data and our synthetic data demonstrate the efficacy of the continuous local features and the advantage of a cross-outfit model over traditional subject-specific ones.

\modelname assumes that the minimally closed body under the clothing is given and the training scans have no noise or missing data.
Dealing with partial, noisy, scans and combining it with automatic body shape and pose estimation~\cite{bhatnagar2020loopreg, PTF:CVPR:2021} are promising future directions for fully-automatic scan animation.
Our use of the UV map for body pose conditioning can sometimes lead to ``seams'' in the outputs (see SupMat.). Future work should explore more continuous parameterizations of the human body manifold.
Additionally, here we factor out dependencies of clothing deformation on body shape for simplicity. 
Given sufficient training data, this would be easy to learn by replacing the neutral shape with the true shape in the UV positional map.

\vspace{3pt}
{\small 
{\bf \noindent Acknowledgements and Disclosure}:
We thank Studio Lupas for helping create the clothing designs, and S.~Saito for the help with data processing during his internship at MPI. 
Q.~Ma is partially funded by DFG (German Research Foundation)-276693517 SFB 1233 and supported by the Max Planck ETH Center for Learning Systems.
The conflict of interest of MJB can be found at {\small \url{https://files.is.tue.mpg.de/black/CoI/ICCV2021.txt}.}
}
\clearpage
\appendix
{\noindent\Large\textbf{Appendix}}
\setcounter{page}{1}
\counterwithin{figure}{section}
\counterwithin{table}{section}

\section{Implementation Details}
\subsection{Model Architecture}
We use the SMPL~\cite{loper2015smpl} (for CAPE data) and SMPL-X~\cite{pavlakos2019expressive} (for \dataset data) UV maps of $128\times128\times3$ resolution as pose input, where each pixel is encoded into 64 channels by the pose encoder. 
The pose encoder is a standard UNet~\cite{ronneberger2015u} that consists of seven [Conv2d, BatchNorm, LeakyReLU(0.2)] blocks, followed by seven [ReLU, ConvTranspose2d, BatchNorm] blocks. The final layer does not apply BatchNorm.

The geometric feature tensor has the same resolution as that of the pose feature tensor, i.e.~$128\times128\times64$. 
It is learned in an auto-decoding~\cite{park2019deepsdf} manner, being treated as a free variable that is optimized together with the network weights during training. 
The geometric feature tensor is followed by three learnable convolutional layers, each with a receptive field of 5, before feeding it to the shape decoder. We find that these convolutional layers help smooth the features spatially, resulting in a lower noise level in the outputs.

The pose and geometric feature tensors are concatenated along the feature channel. 
In all experiments, we query the feature tensor with a $256\times256$ UV map, i.e.~the concatenated feature tensor is spatially $4\times$ bilinearly upsampled. The output point cloud has 50K points.

At each query location, the concatenated pose and geometry feature (64+64-dimensional), together with the 2D UV coordinate of the query point, are fed into an 8-layer MLP. The intermediate layers' dimensions are (256, 256, 256, 386, 256, 256, 256, 3), with a skip connection from the input to the 4th layer as in DeepSDF~\cite{park2019deepsdf}. From the 6th layer, the network branches out 2 heads with the same architecture to predict the displacements and point normals, respectively.
All but the last layer use BatchNorm and a Softplus non-linearity with $\beta=20$. 
The predicted normals are normalized to unit length. 

\subsection{Training} 
We train POP with the Adam~\cite{kingma2014adam} optimizer with a learning rate of $3.0\times10^{-4}$, a batch size of 4, for 400 epochs. 
The displacement and normal prediction modules are trained jointly. 
As the normal loss relies on the nearest neighbor ground truth points found by the Chamfer Distance, we only turn it on when $\mathcal{L}_d$ roughly stabilizes from the 250th epoch. 
The loss weights are set to $\lambda_\textrm{d}=2.0\times10^4,\lambda_\textrm{rd}=2.0\times10^3, \lambda_\textrm{rg}=1.0, \lambda_\textrm{n}=0.0$ at the beginning of the training, and $\lambda_\textrm{n}=0.1$ from the 250th epoch.

\subsection{Data Processing} We normalize all the data examples by removing the body translation and global orientation from them. 
From each clothed body surface, we sample 40K points to serve as training ground truth. 
Note that we do not rely on any connectivity information in the registered meshes from the CAPE dataset.

\subsection{Baselines}
\paragraph{NASA.} We re-implement the NASA~\cite{deng2019neural} model in PyTorch and ensure the performance is on par with that reported in the original paper. 
For evaluating NASA results, we first extract a surface using Marching Cubes~\cite{lorensen1987marching} and then sample the same number of points (50K) from it for a fair comparison. The sampling is performed and averaged over three repetitions.

\paragraph{SCALE.} We employ the same training schedule and the number of patches (798) as in the original SCALE paper~\cite{SCALE:CVPR:2021}, using the implementation released by the authors. We sample 64 points per patch at both training and inference to achieve the same number of output points as ours for a fair comparison.

\paragraph{LBS.} In the main paper Sec.~\ref{exp:scan_animate}, we compare with the Linear Blend Skinning (LBS) in the single scan animation task. This is done with the help of the SMPL~\cite{loper2015smpl} body model: we first optimize the SMPL body shape and pose parameters to fit a minimally-clothed body to the given scan, and then displace the vertices such that the final surface mesh aligns with the scan. The fitted clothed body model is then reposed by the target pose parameters.

\subsection{User Study}
We conduct a large-scale user study on the Amazon Mechanical Turk to get a quantitative evaluation of the visual quality of our model outputs against the point-based method SCALE~\cite{SCALE:CVPR:2021}.
We evaluate over 6,000 unseen test examples in the CAPE and \dataset datasets, from different subjects, performing different poses. 
For each example, the point cloud output from POP and SCALE are both rendered with a surfel-based renderer by Open3D~\cite{Zhou2018open3d} under the same rendering settings (an example of such rendering is the Fig.~\ref{fig:teaser} in the main paper). We then present both images side-by-side to the users and ask them to choose the one that they deem a higher visual quality. 
The left-right ordering of the images is randomly shuffled for each example to avoid users' bias to a certain side. 
The users are required to zoom-in the images before they are able to make the evaluation, and we do not set a time limit for the viewing. 
Each image pair is evaluated by three users, and the final results in the main paper is averaged from all the user choices on all examples.

\section{Datasets}
\begin{figure}
    \centering
    \includegraphics[width=\linewidth]{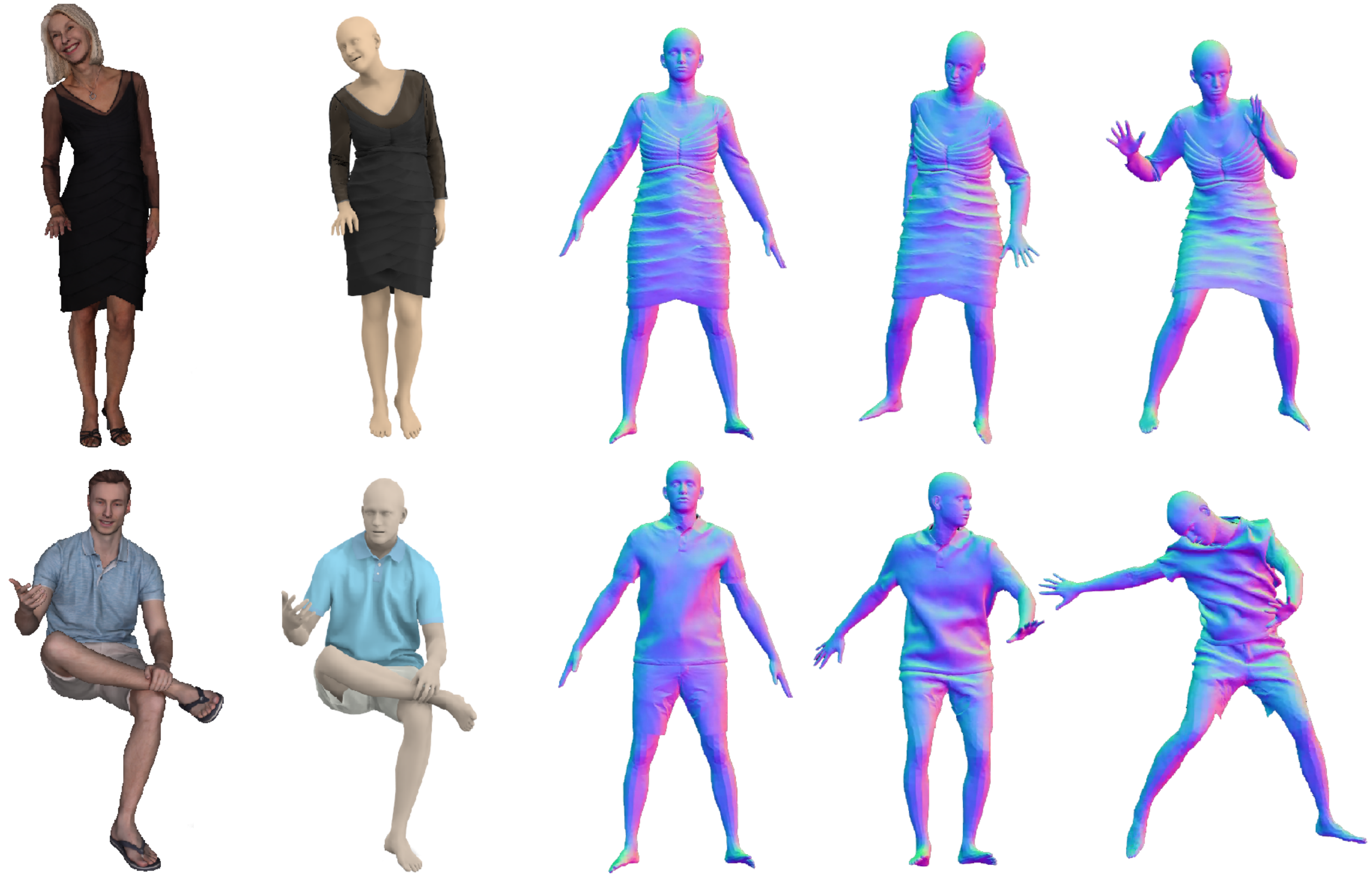}
    \put(-230,-10){\footnotesize{Scans \qquad \quad Designs \qquad \qquad \qquad \qquad Simulations}}
    \caption{Examples from our ReSynth dataset. The clothing is designed based on real-world scans~\cite{renderpeople}, draped on the SMPL-X~\cite{pavlakos2019expressive} body model, and then simulated using Deform Dynamics~\cite{deformdynamics}.}
    \vspace{-15pt}
    \label{fig:ReSynth}
\end{figure}

\paragraph{\dataset.}
The 24 outfits designed by the clothing designer include varied types and styles: shirts, T-shirts, pants, shorts, skirts, long dresses, jackets, to name a few. 
For each outfit, we find the SMPL-X~\cite{pavlakos2019expressive} body (provided by the AGORA~\cite{AGORA:CVPR:21} dataset) that fits the subject's body shape in its corresponding original scan. 
We then use physics-based simulation~\cite{deformdynamics} to drape the clothing on the bodies and animate them with a consistent set of motion sequences of the subject 00096 from the CAPE dataset. The simulation results are inspected manually to remove problematic frames, resulting in 984 frames for training and 347 frames for test for each outfit. 
Examples from ReSynth are shown in Fig.~\ref{fig:ReSynth}. We will release the dataset for research purposes.

\paragraph{CAPE.}
The CAPE dataset~\cite{CAPE:CVPR:20} provides registered mesh pairs of (unclothed body, clothed body) of humans in clothing performing motion sequences. 
The three subjects (00096, 00215, 03375) that we use in the experiments have in total 14 outfits comprising short/long T-shirts, short/long pants, a dress shirt, a polo shirt, and a blazer. For each outfit, the motion sequences are randomly split into training (70\%) and test (30\%) sets. 

\section{Extended Results and Discussions}
Here we provide extended analysis and discussions regarding the main paper Tab.~\ref{tab:exp_summary}. 
The implicit surface baseline, NASA~\cite{deng2019neural}, shows a much higher error than other methods. We find that it is majorly caused by the occasional missing body parts in its predictions. This happens more often for challenging, unseen body poses. The incomplete predictions thus lead to exceptionally high bi-directional Chamfer distance on a number of examples, hence a high average error.

Our approach is based on the SCALE~\cite{SCALE:CVPR:2021} baseline, but it achieves on average 11.4\% (on CAPE data) and 9.1\% (on \dataset) lower errors than SCALE, with both margins being statistically significant ($p$-value$\ll$1$e$-4 in the Wilcoxon signed-rank test). Together with the user study results in the main paper, this shows a consistent improvement on the representation power.

In Figs.~\ref{fig:more_qualitative_cape} and \ref{fig:more_qualitative_rp} we show extended qualitative comparisons with NASA~\cite{deng2019neural} and SCALE~\cite{SCALE:CVPR:2021} from the pose generalization experiment (Sec.~\ref{exp:ours_vs_baselines} in the main paper).
Please refer to the supplementary video at {\small\url{https://qianlim.github.io/POP}} for animated results.

\section{Run-time Comparison}
Here we compare the inference speed of POP with the implicit surface baseline, NASA~\cite{deng2019neural}, and the patch-based baseline, SCALE~\cite{SCALE:CVPR:2021}. 

To generate a point cloud with 50K points, POP takes on average 48.8\textit{ms}, and SCALE takes 42.4\textit{ms}. The optional meshing step using the Poisson Reconstruction~\cite{kazhdan2006poisson} takes 1.1\textit{s} if a mesh is desired.
Both explicit representations have comparable run-time performance. 
In contrast, NASA requires densely evaluating occupancy values over the space in order to reconstruct an explicit surface, which takes 12.2\textit{s} per example.
This shows the speed advantage of the explicit representations over the implicit ones.

\section{Limitations and Failure Cases}
\begin{figure}[ht]
    \centering
    \includegraphics[width=0.95\linewidth]{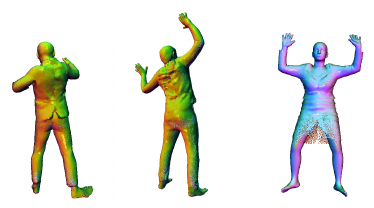}
    \put(-195,-10){\footnotesize{\textbf{(a)}}}
     \put(-125,-10){\footnotesize{\textbf{(b)}}}
      \put(-35,-10){\footnotesize{\textbf{(c)}}}
    \caption{Illustrations of our limitations.}
    \label{fig:failure_case}
\end{figure}

\begin{figure*}[htb]
    \centering
    \includegraphics[width=0.98\linewidth]{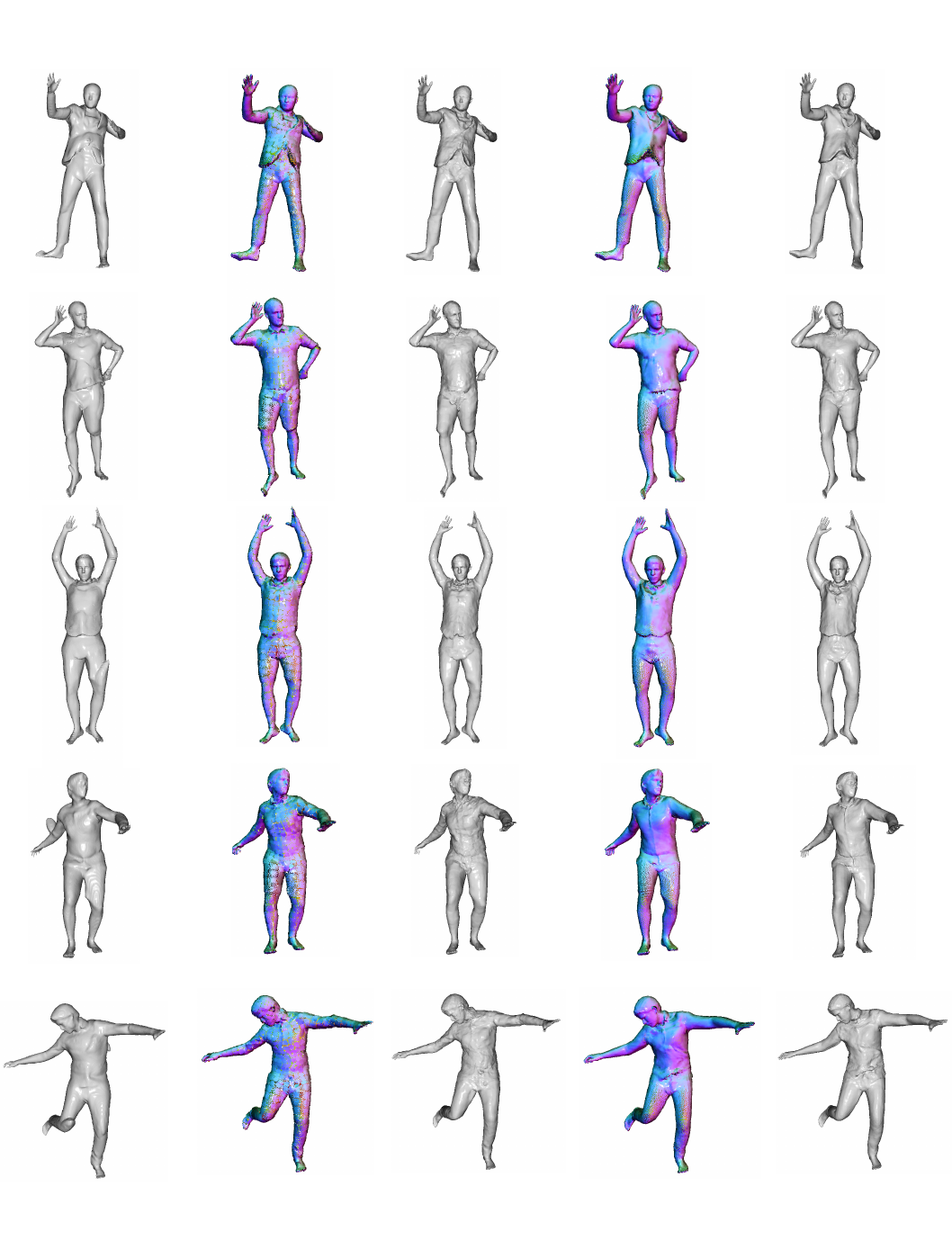}
    \put(-458,620){\footnotesize{NASA~\cite{deng2019neural}}}
    \put(-384,620){\footnotesize{SCALE~\cite{SCALE:CVPR:2021}, point cloud}}
    \put(-287,620){\footnotesize{SCALE~\cite{SCALE:CVPR:2021}, meshed}}
    \put(-188,620){\footnotesize{Ours, point cloud}}
    \put(-83,620){\footnotesize{Ours, meshed}}
    \caption{Extended qualitative results from the pose generalization experiment (main paper Sec.~\ref{exp:ours_vs_baselines}), on the CAPE dataset. Best viewed zoomed-in on a color screen.}
    \label{fig:more_qualitative_cape}
\end{figure*}

\begin{figure*}[htb]
    \centering
    \includegraphics[width=0.92\linewidth]{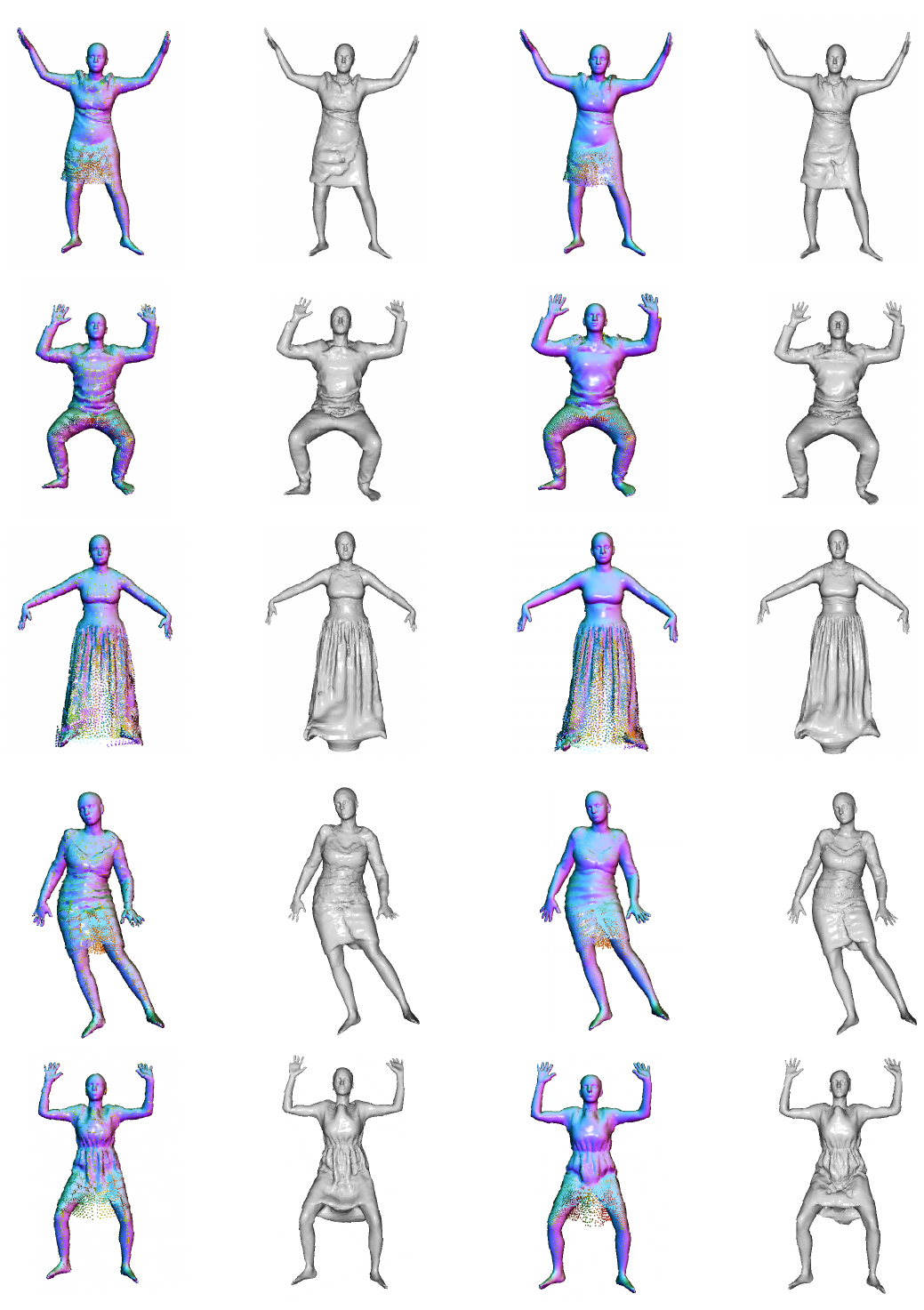}
    \put(-445,650){\footnotesize{SCALE~\cite{SCALE:CVPR:2021}, point cloud}}
    \put(-323,650){\footnotesize{SCALE~\cite{SCALE:CVPR:2021}, meshed}}
    \put(-192,650){\footnotesize{Ours, point cloud}}
    \put(-72,650){\footnotesize{Ours, meshed}}
    \vspace{-10pt}
    \caption{Extended qualitative results from the pose generalization experiment (main paper Sec.~\ref{exp:ours_vs_baselines}), on the \dataset dataset. Best viewed zoomed-in on a color screen.}
    \label{fig:more_qualitative_rp}
\end{figure*}

As discussed in the final section of the main paper, the major limitation of our approach lies on the use of the UV map. 
Although the UV maps are widely used to reconstruct and model human faces~\cite{Thomas_2016_CVPR,feng2018joint,Ma_2021_pixel}, one can encounter additional challenges when applying this technique to human bodies. On a full body UV map such as that of SMPL, different body parts are represented as separate ``islands'' on the UV map, see Fig.~\ref{fig:method_overview} in the main paper. Consequently, the output may suffer from discontinuities at the UV islands' boundaries.
Qualitatively, this may occasionally result in visible ``seams'' between certain body parts as shown in Fig.~\ref{fig:failure_case} (a-b), or an overly sparse distribution of points between the legs in the case of dresses as shown in Fig.~\ref{fig:failure_case} (c), leading to sub-optimal performance when training a unified model for both pants and skirts. 

Note, however, that such discontinuities are not always the case. Intuitively, as the input UV positional map encodes $(x,y,z)$ coordinates on the 3D body, the network can utilize not only the proximity in the UV space but also that in the original 3D space. 
We believe that the problem originates from the simple 2D convolution in the UNet pose encoder. 
A promising solution is to leverage a more continuous parameterization for the body surface manifold that is compatible with existing deep learning architectures. 
We leave this for future work.

\section{Further Discussions on Related Work}
Here we discuss the relationship of our method to recent work that uses similar techniques or that aims similar goals.

We represent clothing as a displacement field on the minimally-clothed body, in the canonical pose space. This helps factor out the effect of the articulated, rigid transformations that are directly accessible from the underlying posed body. In this way, the network can focus on the non-rigid, residual clothing deformation. Such technique is becoming increasingly popular for clothed human shape models that use meshes~\cite{CAPE:CVPR:20}, point clouds~\cite{SCALE:CVPR:2021}, and implicit functions~\cite{palafox2021npm, SCANimate:CVPR:2021, MetaAvatar:arXiv:21, chen2021snarf, liu2021neural,tiwari21neuralgif}. 

Our shape decoder is a coordinate-based multi-layer perceptron (MLP), reminiscent of the recent line of work on neural implicit surfaces~\cite{mescheder2019occupancy, park2019deepsdf,chen2019imnet} and neural radiance fields~\cite{mildenhall2020nerf}. 
These methods learn to map a neural feature at a given query location into a certain quantity, e.g.~a signed distance~\cite{park2019deepsdf}, occupancy~\cite{mescheder2019occupancy}, color and volume density~\cite{mildenhall2020nerf}, or a displacement vector in our case.
Our work differs from others majorly in that the querying coordinates live on a 2-manifold (the body surface)\footnote{The concurrent work by Burov et al.~\cite{burov2021dsfn} also maps the points on the SMPL body surface to a continuous clothing offset field, but the points' Euclidean coordinates (with positional encoding) are used to query the shape decoder.} instead of $\mathbb{R}^3$. 
Moreover, our point cloud representation belongs to the \textit{explicit} representation category, retaining an advantage in the inference speed compared to the implicit methods. With the recent progress in differentiable and efficient surface reconstruction from point clouds~\cite{sharp2020ptn, peng2021sap}, it becomes possible to flexibly convert between point clouds and meshes in various applications.

Recent work on deformable face modeling~\cite{Ma_2021_pixel} and pose-controlled free-view human synthesis~\cite{liu2021neural} employ similar network architectures as ours, despite the difference in the goals, tasks and detailed techniques. While the commonality implies the efficacy of the high-level architectural design, it remains interesting to combine the detailed technical practices from each piece of work. 
It is also interesting to note the connection between our geometric feature tensor and the \textit{neural texture} in recent work on neural rendering~\cite{Raj_2021_anr,thies2019deferred}: both concepts learn a spatial neural representation that controls the output, revealing a connection between modeling the 3D geometry and 2D appearances.

Finally, in concurrent work, MetaAvatar~\cite{MetaAvatar:arXiv:21} also learns a multi-subject model of clothed humans, which can generalize to unseen subjects using only a few depth images. Unlike our auto-decoding learning of the geometric feature tensor, MetaAvatar uses meta-learning to learn a prior of pose-dependent cloth deformation across different subjects and clothing types that helps generalize to unseen subjects and clothing types at test-time.
We believe both approaches will inspire future work on cross-garment modeling and automatic avatar creation.

{\small
\bibliographystyle{ieee_fullname}
\bibliography{references}
}

\end{document}